\documentclass[conference]{IEEEtran}
\IEEEoverridecommandlockouts
\usepackage{cite}
\usepackage{amsmath,amssymb,amsfonts}
\usepackage{graphicx}
\usepackage{diagbox}
\usepackage{booktabs}
\usepackage{subfigure}
\usepackage{textcomp}
\usepackage{xcolor}
\usepackage{bbm}
\usepackage{multirow}
\usepackage{algorithm}
\usepackage{algorithmic}
\usepackage{algorithm}
\usepackage{tabularray}
\DeclareMathOperator*{\sign}{sign}
\def\BibTeX{{\rm B\kern-.05em{\sc i\kern-.025em b}\kern-.08em
    T\kern-.1667em\lower.7ex\hbox{E}\kern-.125emX}}
\begin{document}

\title{Towards Robust Federated Learning via Logits Calibration on Non-IID Data \\
}

\author{\IEEEauthorblockN{Yu Qiao\textsuperscript{1}, Apurba Adhikary\textsuperscript{2}, Chaoning Zhang\textsuperscript{1}, and Choong Seon Hong\textsuperscript{2}}
\IEEEauthorblockA{
\textsuperscript{1}
\textit{Department of Artificial Intelligence, Kyung Hee University, Yongin-si 17104, Republic of Korea}\\
\textsuperscript{2}
\textit{Department of Computer Science and Engineering, Kyung Hee University, Yongin-si 17104, Republic of Korea}\\
E-mail: \ qiaoyu@khu.ac.kr, apurba@khu.ac.kr, chaoningzhang1990@gmail.com, cshong@khu.ac.kr}}

\maketitle

\begin{abstract}
Federated learning (FL) is a privacy-preserving distributed management framework based on collaborative model training of distributed devices in edge networks. However, recent studies have shown that FL is vulnerable to adversarial examples (AEs), leading to a significant drop in its performance. Meanwhile, the non-independent and identically distributed (non-IID) challenge of data distribution between edge devices can further degrade the performance of models. Consequently, both AEs and non-IID pose challenges to deploying robust learning models at the edge. In this work, we adopt the adversarial training (AT) framework to improve the robustness of FL models against adversarial example (AE) attacks, which can be termed as federated adversarial training (FAT). Moreover, we address the non-IID challenge by implementing a simple yet effective logits calibration strategy under the FAT framework, which can enhance the robustness of models when subjected to adversarial attacks. Specifically, we employ a direct strategy to adjust the logits output by assigning higher weights to classes with small samples during training. This approach effectively tackles the class imbalance in the training data, with the goal of mitigating biases between local and global models. Experimental results on three dataset benchmarks, MNIST, Fashion-MNIST, and CIFAR-10 show that our strategy achieves competitive results in natural and robust accuracy compared to several baselines.
\end{abstract}

\section{Introduction}
\label{sec:intro}
Mobile edge computing (MEC) has introduced a paradigm shift in distributed computing by moving computational resources closer to data sources and edge devices~\cite{wang2023wireless}. This advancement has paved the way for artificial intelligence (AI)-based~\cite{adhikary2023artificial_noms,raha2023segment,raha2023generative} approaches such as federated learning (FL)~\cite{mcmahan2017communication} for distributed network management, which leverages the cooperation between edge devices and server for global model training and inference in a distributed manner. By combining the power of edge computing with the privacy-preserving natural of FL, it becomes possible to allow edge devices to collaboratively train a shared global model without exposing their raw data, making it a promising approach for edge intelligence~\cite{mcmahan2017communication,qiao2023mp,zhu2021federated}. However, there are non-independent and identically distributed (non-IID) concerns between devices in FL, which may cause inconsistencies in the update directions of the local model and the global model, and eventually lead to model training failure to converge~\cite{qiao2023cdfed,li2020federated}. In addition, FL is also found to be vulnerable to adversarial attacks in recent studies~\cite{zizzo2020fat,hong2021federated,lyu2022privacy}. Specifically, attackers can fool the model into making wrong decisions by crafting adversarial examples (AEs) that are imperceptible to humans during the model inference phase. Given such security issues as well as the challenge of non-IID, it is necessary to design a robust FL paradigm to resist adversarial attacks in non-IID settings.

Adversarial training (AT)~\cite{goodfellow2014explaining} has been considered as one of the most effective defense mechanisms to defend against adversarial attacks in traditional centralized machine learning~\cite{shafahi2019adversarial}. Essentially, this approach directly allows the model to learn based on AEs, so that the trained model is robust to adversarial perturbations. AT was formulated as a minimax optimization problem, and projected gradient descent (PGD) was a suggested algorithm to improve the robustness of models against adversarial attacks in AT~\cite{madry2017towards}. Recently, this training paradigm has also been shown to be promising in enhancing the adversarial robustness of the global model by performing local AT at the edge of FL, which is termed as federated adversarial training (FAT)~\cite{zizzo2020fat,zhang2023delving}. Note that the robustness of the model to adversarial attacks inevitably leads to a loss of accuracy compared to natural training (i.e. model training on clean samples without perturbation)~\cite{chen2022calfat}. Subsequently, \cite{shah2021adversarial} explored the possibility of performing FAT with limited communication resources. However, these methods ignore the natural accuracy of FAT with clean samples. On the other hand, there are several works~\cite{mcmahan2017communication, qiao2023mp, qiao2023framework, li2020federated, li2021model} that aim to deal with non-IID issues in FL. FedAvg~\cite{mcmahan2017communication} is the first algorithm to deal with non-IID challenges in FL. It addresses this challenge through a three-step training process: 1) The server distributes the global model to participating edge devices. 2) Each edge device updates the global model based on its local data and sends the updated model back to the server. 3) The server aggregates the model parameters from the devices using a weighted average strategy and sends the updated global model back to the edge devices. This iterative process continues until convergence. For example, ~\cite{li2020federated,li2021model} and~\cite{qiao2023framework, qiao2023mp} respectively limit the inconsistency of the update direction of local models and the global model from the perspectives of regularization and prototype~\cite{qiao2023boosting}. However, none of these studies investigated the robustness issue in FL, which poses a risk to model deployment at the edge.

This work aims to improve the accuracy and robustness of FL models against adversarial attacks in non-IID settings. For the non-IID setting, we focus on the case where the label distribution of the device data is non-IID, i.e. the number of samples with different labels is unbalanced within each device. The unbalanced sample size will cause the prediction results of local models to be biased towards those classes with a large number of samples, resulting in biased prediction results~\cite{zhang2022federated}. Inspired by the class frequency idea~\cite{menon2020long}, we attempt to eliminate the bias within each device by calibrating the logit output for each class. The logits denote the output of the classification layer and serve as the input to the softmax function. Specifically, in each local epoch, we count the occurrence frequency of different classes in each mini-batch iteration for each client. We then use a direct method to find the square root of the inverse frequency of each class, that is, set the logit weight of each class output to the reciprocal of the square root of the frequency of occurrence in the mini-batch. The inverse frequency can help the model pay more attention to rare classes by giving higher weight to them, thus improving the performance on imbalanced classes. In other words, this strategy can balance the weight difference between rare classes and common classes to some extent, reducing the impact of rare classes on model training while maintaining relatively high weight values. The main contributions are summarized as follows:
\begin{itemize}
\item Based on the AT strategy, we introduce an adversarial robust federated management framework through logits calibration, termed FedALC, which can improve the robustness under non-IID challenges.
\item We propose to improve the robustness of the model under varying degrees of data heterogeneity by exploiting logits calibration and AT strategy. Specifically, we calibrate the local AT process by incorporating square roots of the inverse frequencies of different classes as weights for logits adjustment in each local epoch. 
\item We conduct experiments over three prevalent datasets: MNIST\cite{lecun1998gradient}, Fashion-MNIST\cite{xiao2017fashion}, and CIFAR-10~\cite{krizhevsky2009learning}. The results show that our proposal has a competitive performance gain and a faster convergence rate than several baselines.
\end{itemize}

The remaining sections of this paper are structured as follows. Section \ref{sec:preliminary} outlines the main training process of standard FL and FAT. The methodology is detailed in Section \ref{sec:method}. Section \ref{sec:exps} presents comprehensive experimental results. Finally, in Section \ref{sec:cons}, we conclude the work.

\begin{figure}[t]
\centering
\includegraphics[width=0.50\textwidth]{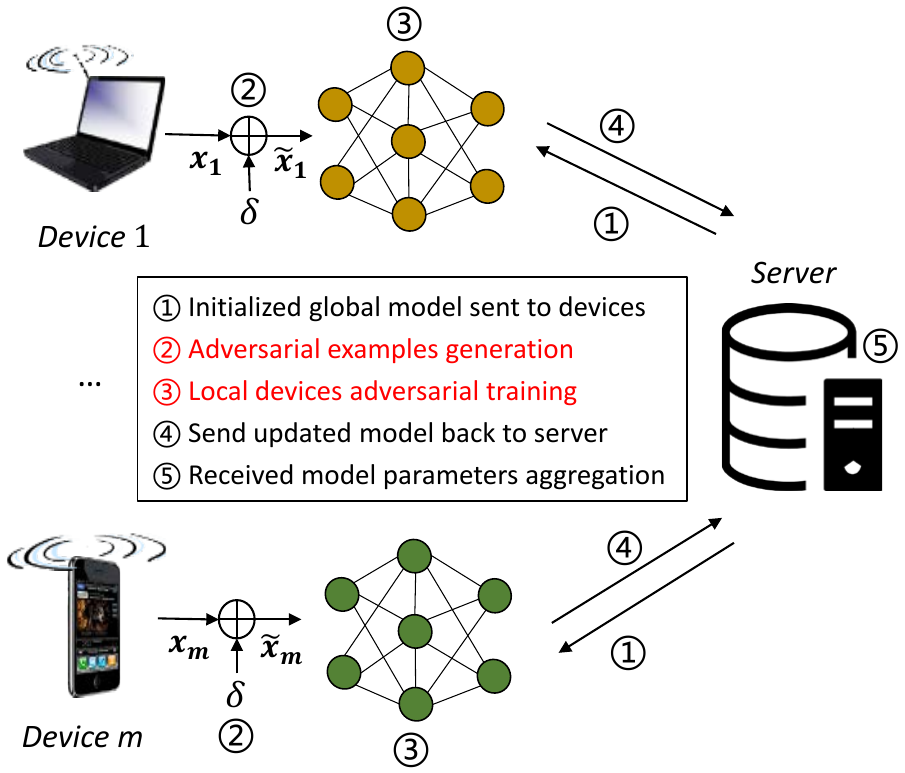}
\caption{The overview of the proposed FAT process. The main difference from the standard FL is mainly in the local training phase of the device, where FAT decomposes the local training phase of the device in the standard FL into step 2 and step 3 in the illustration. In this work, we focus on step 2 and step 3.}
\label{fig:System_model}
\end{figure}

\section{Preliminaries}
\label{sec:preliminary}
\subsection{Standard Federated Learning}
We consider $M$ edge devices and a single server for the environment of federated management, where each device $i$ holds its private and sensitive dataset $\mathcal{D}_i$ consisting of feature and label pairs denoted as $\boldsymbol{x}_i$ and $y_i$, respectively. $D_i$ represents the dataset size owned by each device. The local loss for device $i$ can be expressed as follows ~\cite{qiao2023framework}:
\begin{equation} \label{eq:one_hot}
   {\mathcal{L}_i} (\theta_i)= -\frac {1}{D_i} \sum_{i \in \mathcal{D}_i}\sum_{j=1}^{C}\mathbbm{1}_{y = j}\log \frac{exp(z_j)}{exp(\sum_{j=1}^{C}{z_j})},
\end{equation}
where $z$ represents the logit value of the local model $\mathcal F(\theta; \boldsymbol{x}_i)$ for each device. $\mathbbm{1}(\cdot)$ is the indicator function. The label space $[C]$ contains $C$ classes, where $C$ represents the number of classes. $\theta_i$ represents the model parameters of each local device.

Therefore, the global objective of standard FL can be represented as follows, which minimizes the sum of local losses across distributed devices.
\begin{equation} \label{global_loss}
    \mathop {\min}_{\theta} {\mathcal{L}} (\theta) = \sum_{i \in [M]} \frac {D_i}{\sum_{i \in [M]}D_i} {\mathcal{L}_i} (\theta_i),
\end{equation}
where $[M]$ denotes the set of distributed devices with $[M] = \{1,...,M\}$.

\subsection{Federated Adversarial Learning}
\label{sec:fat}
The main component of FAT is to introduce imperceptible perturbations to the training samples during the standard federated training phase, allowing the model to make correct predictions even in the presence of these perturbations. The AT adopted by FAT is an optimization problem, which usually uses the PGD algorithm to generate AEs in each iteration. Subsequently, model parameters are optimized to minimize the impact of these AEs on the model, thereby enhancing the robustness of the model to adversarial perturbations. Therefore, according to the formula in~\cite{madry2017towards}, Eq.~\ref{global_loss} can be rewritten as follows:

\begin{equation} \label{eq:adv_gener}
    \mathop {\min}_{\theta} {\mathcal{L}_{adv}} (\theta) = \sum_{i \in [M]} \frac {D_i}{\sum_{i \in [M]}D_i} {\mathcal{L}^{adv}_i} (\theta_i),
\end{equation}
where $\mathcal{L}_{adv}(\theta_i)$ and $\mathcal{L}^{adv}_i(\theta)$ are local loss and global loss obtained through AT, respectively. $\mathcal{L}^{adv}_i(\theta_i)$ = $\mathbb{E}_{(\boldsymbol {x}_i; y_i) \in \mathcal{D}_i}[\mathop {max} \mathcal{L}^{adv}_i(\mathcal F(\theta_i; \widetilde {\boldsymbol {x}_i}); y_i))]$, and the $\widetilde {\boldsymbol {x}_i}$ is the AE of $\boldsymbol {x}_i$, which is generated by PGD as follows:
\begin{equation}
\label{eq:pgd}
\boldsymbol{x_i}^{t+1}=\Pi_{\boldsymbol{x_i} + \delta}\left(\boldsymbol{x_i}^t+\alpha\sign(\nabla_{\boldsymbol x}  \mathcal{L}_i(\mathcal F(\theta_i; \boldsymbol {x}_i); y_i))) \right),
\end{equation}
where $\alpha$ denotes the step size, $\boldsymbol{x_i}^t$ marks the AE generated at step $t$, $\Pi_{\boldsymbol{x_i} + \delta}$ represents the projection function that projects the AE onto the $\epsilon$-ball centered at $\boldsymbol{x_i}^0$, and $\sign(\cdot$) indicates the sign function. Further, to confirm that the perturbation $\delta$ is imperceptible (or quasi-imperceptible) to the human eye, it is typically constrained by an upper bound $\epsilon$ on the $\ell_\infty$-norm, i.e., $||\delta||_\infty \leq \epsilon$.

The concept of FAT was initially introduced by~\cite{zizzo2020fat} as a solution to resist the vulnerability of FL on adversarial examples. However, it does not take into account the accuracy under clean samples, and only applies AT to a subset of data of the local device to enhance the robustness of the global model. Furthermore, using only a portion of the data for AT may limit the exploitation of the diversity and complexity of the entire dataset, limiting the potential for improving model robustness. We follow the strategy described in~\cite{zizzo2020fat}, which involves conducting AT on local models to enhance their robustness and consequently improve the overall robustness of the global model. However, in order to make full use of each client's local data, we redefine the procedure of local training for each device in FAT, whereby each client can use its own complete dataset to participate in adversarial training. Specifically, the key difference from standard FL lies in the second step, known as local training at the edge sides. In this step, each client is tasked with generating AEs using their own data during the local epochs. Subsequently, the clients update their respective model parameters based on the generated adversarial samples and ground truth labels, thereby enhancing the robustness of their local models against adversarial attacks.

\section{Methodology}
\label{sec:method}
\subsection{Proposed FAT Framework}
We illustrate the main training process of federated adversarial learning, termed FedALC, as shown in Figure \ref{fig:System_model}. For simplicity, we only mark one global iteration round in this figure. Similar to the standard FL process, the server first sends the initial model parameters to the participating edge devices for individual local updates (i.e. initialized global model sent to devices in step 1). Next, each local device updates the received model parameters based on its own local dataset (i.e., local training phase in step 2 and step 3. Subsequently, all participating devices send their updated model parameters back to the server for aggregation (i.e., send updated model back to server in step 4). Finally, all the received model parameters are aggregated at the server and start the next global iteration, repeating these steps until convergence. The main difference from standard FL is that in the local training phase (i.e, step 2 and step 3), each local model needs to be trained against disturbance $\delta$. Specifically, an invisible perturbation $\delta$ needs to be added to the respective input data $\boldsymbol {x}_i$ during the local training phase of each device (i.e., generate adversarial samples $\widetilde{\boldsymbol {x}}_i$ in step 2), and then each device need to update the respective local model parameters to resist this adversarial perturbation (i.e., local devices adversarial training in step 3). 

\subsection{Calibrated Local Adversarial Training Phase}
\label{sec:cali_local_training}
In the deep learning paradigm, neural network architectures typically consist of three main components: the input layer, the hidden layers, and the classification layer. The hidden layers are responsible for mapping the input space to an embedded space, while the classification layer is responsible for mapping the embedded space to a logit space. The predicted logits serve as the inputs to the computation of a certain loss, which measures the dissimilarity between the predicted logits and the true labels. By comparing the predicted logits with the ground truth labels, the model parameters are iteratively updated to minimize the loss and improve the accuracy of the model. However, as mentioned earlier, calculating the loss directly using the logits can lead to a bias in the local model updates towards classes with larger sample sizes~\cite{zhang2022federated}. In this subsection, we calibrate the logits of each class before feeding them into the loss function (note that this work adopts cross-entropy loss) to mitigate the bias of local updates. By calibrating the output logits, we aim to assign higher importance to the minority class samples, allowing the model to pay more attention to those samples during the training process. This helps in reducing the impact of class imbalance and ensures that the model learns equally well across all classes, regardless of their distribution in the training data.

Let $C_i$ be the set of classes for each device. For the $i$-th device, let $n_j$ represent the number of samples of $j$-th class within the current batch, and $N_i$ represent the number of samples within each batch for each device. Within a batch, the calculation of the square root of the frequency for each class in the set $C_i$ can be expressed as follows:

\begin{equation} \label{eq:frequency}
 w_{i,j}  = \frac{n_j}{sqrt(N_i)},  \quad j \in C_i,
\end{equation}
where $w_{i,j}$ denotes the weight of $j$-th class for each device $i$ within a mini-batch samples. Note that, in our experimental setup, we empirically adopt the square root in this re-weighting formula. The choice for a more straightforward and intuitive approach without the square root is left for future work.

To calibrate the output of the classification layer (a.k.a logits value), we take $w_{i,j}$ as the weight of the corresponding logits value. Specifically, we denote the output of the classification layer as $z^{adv}_{i,j}(\theta_{i}; \widetilde{\boldsymbol{x}}_{i,j})$, where $\widetilde{\boldsymbol{x}}_{i,j}$ represents AE of input data belonging to class $j$ for each device. 

\begin{equation} \label{eq:cali_logit}
\hat{z}^{adv}_{i,j}  = w_{i,j} \cdot z^{adv}_{i,j}(\theta_{i}; \widetilde{\boldsymbol{x}}_{i,j}) ,  \quad j \in C_i,
\end{equation}
where $\hat{z}^{adv}_{i,j}$ represents the weighted logits of $j$-th class for each device. The Eq.~\ref{eq:one_hot} can then be rewritten as follows:
\begin{equation} \label{eq:one_hot_weight}
   {\mathcal{L}^{cali}_i} (\theta_i)= -\frac {1}{D_i} \sum_{i \in \mathcal{D}_i}\sum_{j=1}^{C}\mathbbm{1}_{y = j}\log \frac{exp(\hat{z}^{adv}_j)}{exp(\sum_{j=1}^{C}{\hat{z}^{adv}_j})}.
\end{equation}

Therefore, the global objective of adversarial FL can be reformulated as follows, which aims to minimize the sum of local losses across distributed devices after AT.
\begin{equation} \label{global_loss2}
    \mathop {\min}_{\theta} {\mathcal{L}_{adv}^{cali}} (\theta) = \sum_{i \in [M]} \frac {D_i}{\sum_{i \in [M]}D_i} {\mathcal{L}^{cali}_i} (\theta_i).
\end{equation}

Further details about the FAT process are outlined in Algorithm \ref{alg:FedALC}. In Algorithm \ref{alg:FedALC}, the initialization of the algorithm is performed as specified in line 1. The global rounds process is handled in lines 2 to 8, while lines 9 to 17 are responsible for calculating the local model updates for each edge device. In each global round, the updated global model parameters are distributed to each device in line 4. Each device then updates its received model based on its own dataset $\mathcal{D}_i$ in line 5. For each device, after generating adversarial examples in line 11, it proceeds to calculate its calibrated logit value, and cross-entropy loss in lines 12, and 13, respectively. Based on these calculations, the device updates its model using gradient descent in line 14. The server handles the received updated model parameters in line 7. The algorithm continues to execute for a specified number of global rounds $T$ until convergence.

\begin{algorithm}[t] 
    \caption{FedALC} 
    \label{alg:FedALC} 
    \begin{algorithmic}[1]
        \REQUIRE ~~ \\
        Dataset $\mathcal{D}_i$ of each device, $\theta_i$,
        \STATE \textbf{Initialize $\theta^0$}.
        \FOR{ $t$ = 1, 2, ..., $T$} 
            \FOR{ $i$ = 0, 1,..., $M$ \textbf{in parallel}}
                \STATE Send global model $\theta^t$ to device \textit{i}
                \STATE {$\theta_i^t \gets \textbf{LocalUpdate}(\theta_i^t$)}
            \ENDFOR
            \STATE {$\theta^{t+1} \gets \sum_{i=1}^{M} \frac {D_i}{\sum_{i=1}^{M} D_i}\theta_i^t$}
        \ENDFOR \\
        \textbf {LocalUpdate($\theta_i^t$)}
        \FOR{ each local epoch }
        \FOR{ each batch ($\boldsymbol{x}_i$; $y_i$) of $\mathcal{D}_i$ }
        \STATE {Adversarial examples generation in Eq.~\ref{eq:adv_gener}} 
        \STATE {Calibrate logit value for each device using Eq.~\ref{eq:cali_logit}} 
        \STATE Local model updates using calibrated loss in Eq.~\ref{eq:one_hot_weight}
        \STATE {$\theta_i^t \gets \theta_i^t - \eta\nabla \mathcal L^{cali}_{i}$}
    \ENDFOR
    \ENDFOR
        \RETURN $\theta_i^t$
    \end{algorithmic}
\end{algorithm}

\section{Experiments}
\label{sec:exps}
\subsection{Implementation Details}
\textbf{Datasets and local models.}
We compare the performance of FedALC with several baselines, including FedAvg~\cite{mcmahan2017communication} and FedProx~\cite{li2020federated}, by conducting experiments on three widely used benchmark datasets: MNIST~\cite{lecun1998gradient}, Fashion-MNIST~\cite{xiao2017fashion}, and CIFAR-10~\cite{krizhevsky2009learning}. MNIST is a dataset of handwritten digits containing 10 categories, while Fashion-MNIST contains a total of 10 categories, including different types of clothing and accessories. CIFAR-10 is a more complex dataset, where each sample is a 32x32x3 color image, consisting of 10 classes, with 6,000 images per class, for a total of 50,000 training samples and 10,000 testing samples. We adopt the same model architecture to fairly compare all baselines. Specifically, we employ a multi-layer CNN with 2 convolutional layers and 2 fully connected layers for MNIST and Fashion-MNIST, while for CIFAR-10, we adopt ResNet-18~\cite{he2016deep} pre-trained on ImageNet.

\textbf{Hyperparameters and metrics.} 
Following previous work~\cite{qiao2023cdfed,qiao2023mp,mu2023fedproc}, all baselines follow the Dirichlet distribution Dir($\alpha$) to set the non-IID. Since MNIST and Fashion-MNIST need to be trained from scratch, we set the number of communication rounds to 100, while for pre-trained CIFAR-10, we only set the number of communication rounds to 60. Here, a smaller value of $\alpha$ indicates a larger skewness of the data distribution among devices, and vice versa. Note that we select 5,000 random samples to conduct experiments on MNIST and Fashion-MNIST, and 1,000 random samples on CIFAR-10, since our goal is to evaluate the effectiveness of the proposed method. Furthermore, we use the Adam optimizer~\cite{kingma2014adam} and set the number of clients, local batch size, and learning rate as 10, 32, and 0.001, respectively. In particular, since Fashion-MNIST and CIFAR-10 are more complex than MNIST, we set the local epochs to 5 for the former, and we set 1 for the latter. For evaluation, we report natural test accuracy (i.e., samples without adversarial perturbations) and robust test accuracy under adversarial perturbations. The adversarial data is generated by Fast Gradient Sign Method (FGSM)~\cite{goodfellow2014explaining}, Projected Gradient Descent (PGD)~\cite{madry2017towards}, Basic Iterative Method (BIM)~\cite{kurakin2018adversarial}, and CW~\cite{carlini2017towards}. We set the perturbation bound $\delta$ to be $8/255$ for FGSM, BIM, and PGD attacks. BIM, PGD, and FGSM have a step size of $2/255$.

\begin{table*}[t]
\caption{Natural and robust accuracy (\%) on different datasets. The best results are highlighted in \textbf{bold}.
}
\label{tbl:dataset}
\centering
\resizebox{\textwidth}{!}{
\begin{tabular}{c|ccccc|ccccc}
\toprule
Dataset & \multicolumn{5}{c|}{MNIST} & \multicolumn{5}{c}{Fashion-MNIST} \\
\midrule
$\alpha=0.05$ & Natural & FGSM & BIM & CW & PGD & Natural & FGSM & BIM & CW & PGD \\
\midrule
FedAvg & 
83.84 $\pm$ 0.22 & 40.15 $\pm$ 0.21 & 39.83 $\pm$ 0.16 & 41.44 $\pm$ 0.16 & 39.96 $\pm$ 0.21 & 

69.85 $\pm$ 0.39 & 29.91 $\pm$ 0.45 & 29.41 $\pm$ 0.28 & 35.64 $\pm$ 0.20 & 28.70 $\pm$ 0.18 \\

FedProx & 
80.12 $\pm$ 0.31 & 38.25 $\pm$ 0.14 & 37.95 $\pm$ 0.20 & 38.58 $\pm$ 0.40 & 38.13 $\pm$ 0.19 & 

66.30 $\pm$ 0.64 & 29.66	$\pm$ 0.15 & 28.76 $\pm$ 0.23 & 32.37 $\pm$ 0.26 & 27.98 $\pm$ 0.17 \\

FedALC (ours) & 
\textbf{85.62} $\pm$ 0.21 & \textbf{40.76} $\pm$ 0.19 & \textbf{40.65} $\pm$ 0.22 & \textbf{42.42} $\pm$ 0.37 & \textbf{40.43} $\pm$ 0.26 &

\textbf{71.84} $\pm$ 0.56 & \textbf{30.80} $\pm$ 0.15 & \textbf{29.73} $\pm$ 0.19 & \textbf{36.79} $\pm$ 0.22 & \textbf{30.18} $\pm$ 0.14 \\
\midrule
$\alpha=0.1$ & Natural & FGSM & BIM & CW & PGD & Natural & FGSM & BIM & CW & PGD \\
\midrule
FedAvg & 
79.51 $\pm$ 0.43 & 38.56 $\pm$ 0.20 & 38.24 $\pm$ 0.27 & 39.54 $\pm$ 0.26 & 38.57 $\pm$ 0.25 & 

69.94 $\pm$ 0.44 & 30.33 $\pm$ 0.26 & 29.62 $\pm$ 0.12 & 34.68 $\pm$ 0.22 & 29.31	$\pm$ 0.16 \\

FedProx & 
71.70 $\pm$ 0.25 & 33.74 $\pm$ 0.25 & 33.21 $\pm$ 0.36 & 36.67 $\pm$ 0.22 & 33.69	$\pm$ 0.29 & 
63.81 $\pm$ 0.72 & 29.17 $\pm$ 0.10 & 28.01 $\pm$ 0.16 & 31.99 $\pm$ 0.24 & 28.28	$\pm$ 0.13 \\

FedALC (ours) & 
\textbf{82.62} $\pm$ 0.35 & \textbf{40.00} $\pm$ 0.20 & \textbf{40.02} $\pm$ 0.17 & \textbf{41.41} $\pm$ 0.20 & \textbf{40.11} $\pm$ 0.26 & 

\textbf{72.38} $\pm$ 0.55 & \textbf{31.84} $\pm$ 0.13 & \textbf{31.24} $\pm$ 0.11 & \textbf{36.92} $\pm$ 0.15 & \textbf{30.78} $\pm$ 0.14 \\
\midrule
$\alpha=0.5$ & Natural & FGSM & BIM & CW & PGD & Natural & FGSM & BIM & CW & PGD \\
\midrule
FedAvg & 
95.35 $\pm$ 0.09 & 46.84 $\pm$ 0.07 & 46.73 $\pm$ 0.05 & 47.64 $\pm$ 0.06 & 46.72 $\pm$ 0.05 & 

\textbf{84.27} $\pm$ 0.05 & 37.67 $\pm$ 0.11 & 36.65	$\pm$ 0.07 & \textbf{42.14} $\pm$ 0.06 & 36.47 $\pm$ 0.07 \\
FedProx & 
92.14 $\pm$ 0.19 & 45.42 $\pm$ 0.14 & 45.27 $\pm$ 0.11 & 46.82 $\pm$ 0.12 & 45.35 $\pm$ 0.14 & 
82.82 $\pm$ 0.35 & 36.93 $\pm$ 0.05 & 35.56 $\pm$ 0.11 & 39.96 $\pm$ 0.17 & 36.29 $\pm$ 0.06 \\
FedALC (ours) & 
\textbf{95.90} $\pm$ 0.11 & \textbf{47.06} $\pm$ 0.03 & \textbf{46.91} $\pm$ 0.05 & \textbf{47.98} $\pm$ 0.04 & \textbf{46.91} $\pm$ 0.06 & 

84.21 $\pm$ 0.10 & \textbf{38.07} $\pm$ 0.08 & \textbf{36.80} $\pm$ 0.05 & 41.76 $\pm$ 0.05 & \textbf{37.22} $\pm$ 0.10 \\
\midrule

$\alpha=1.0$ & Natural & FGSM & BIM & CW & PGD & Natural & FGSM & BIM & CW & PGD \\
\midrule
FedAvg & 
96.00 $\pm$ 0.10 & 47.36 $\pm$ 0.03 & 47.16 $\pm$ 0.05 & 47.74 $\pm$ 0.05 & 47.15 $\pm$ 0.04 & 
84.96 $\pm$ 0.09 & 38.21 $\pm$ 0.09 & 37.01 $\pm$ 0.07 & 42.28 $\pm$ 0.03 & 37.25 $\pm$ 0.07 \\
FedProx & 
94.27 $\pm$ 0.12 & 46.63 $\pm$ 0.05 & 46.32 $\pm$ 0.06 & 47.63 $\pm$ 0.03 & 46.32 $\pm$ 0.06 & 
84.39 $\pm$ 0.10 & 37.52 $\pm$ 0.09 & 36.66 $\pm$ 0.03 & 42.14 $\pm$ 0.05 & 36.53 $\pm$ 0.08 \\
FedALC (ours) & 
\textbf{96.40} $\pm$ 0.02 & \textbf{47.58} $\pm$ 0.05 & \textbf{47.34} $\pm$ 0.03 & \textbf{48.30} $\pm$ 0.03 & \textbf{47.45} $\pm$ 0.06 &
\textbf{85.29} $\pm$ 0.11 & \textbf{38.76} $\pm$ 0.09 & \textbf{37.52} $\pm$ 0.06 & \textbf{42.70} $\pm$ 0.03 & \textbf{37.69} $\pm$ 0.10 \\
\bottomrule
\end{tabular}}
\end{table*}

\begin{figure}[!htp]
\centering
\subfigure{\includegraphics[width=0.39\textwidth]{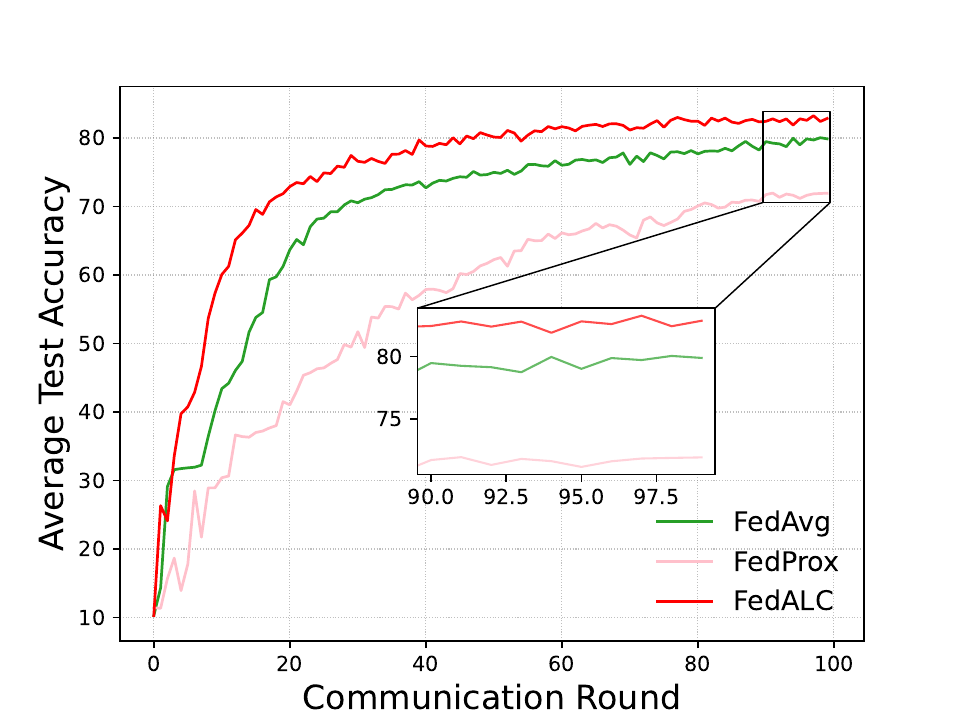}}
\subfigure{\includegraphics[width=0.39\textwidth]{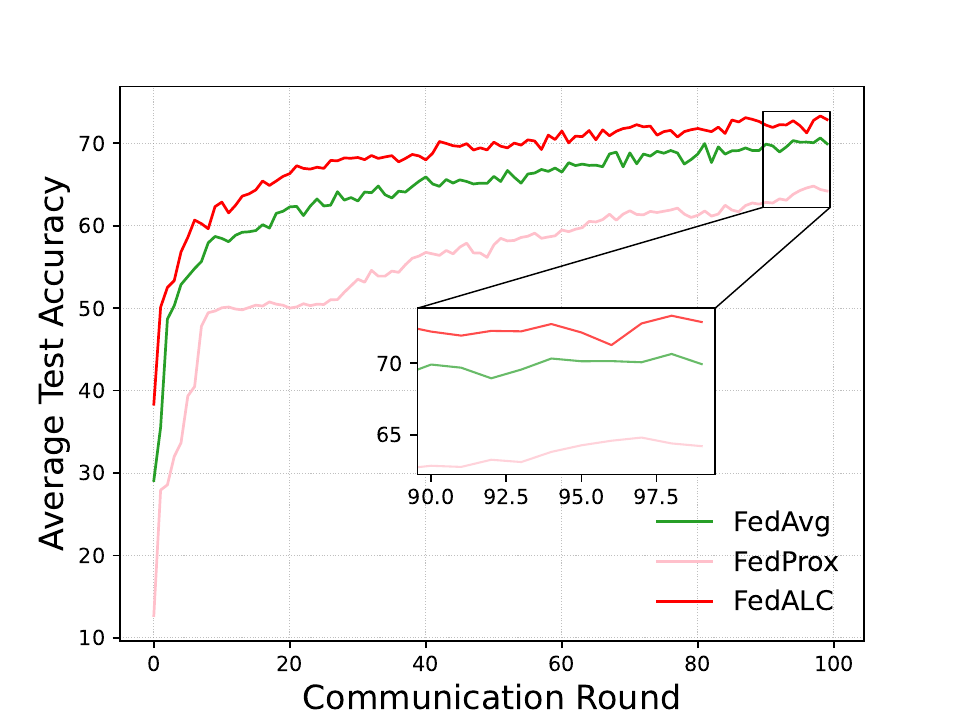}}
\caption{The top-1 average natural test accuracy (\%) of all methods on MNIST (top) and Fashion-MNIST (down) with Dir(0.1) in each global round.}
\label{comm_effici_natural}
\end{figure}

\begin{figure}[!htp]
\centering
\subfigure{\includegraphics[width=0.39\textwidth]{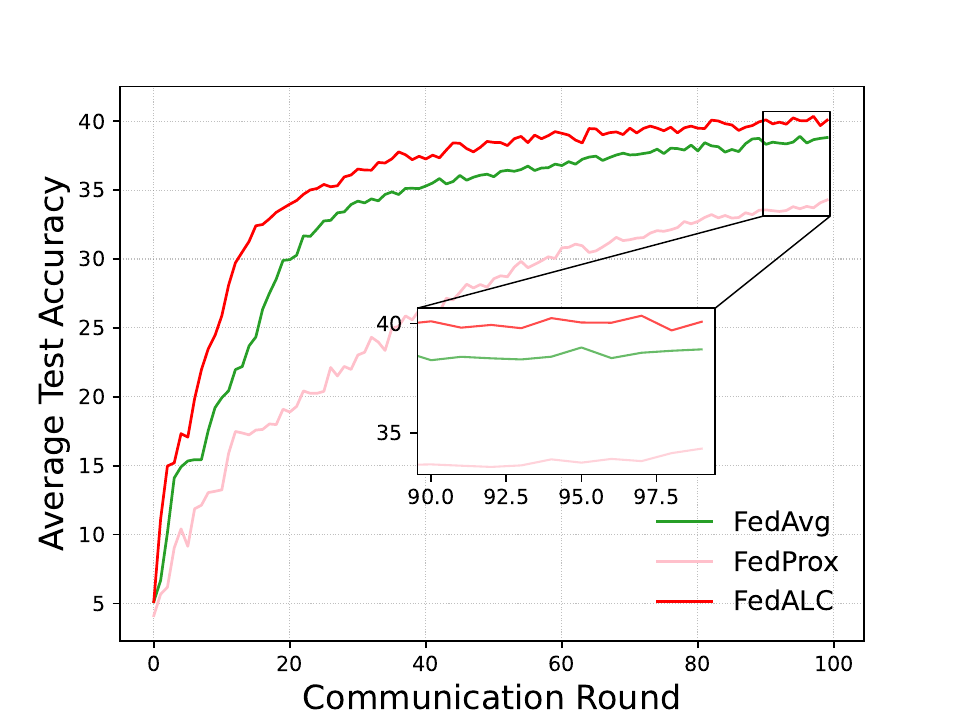}}
\subfigure{\includegraphics[width=0.39\textwidth]{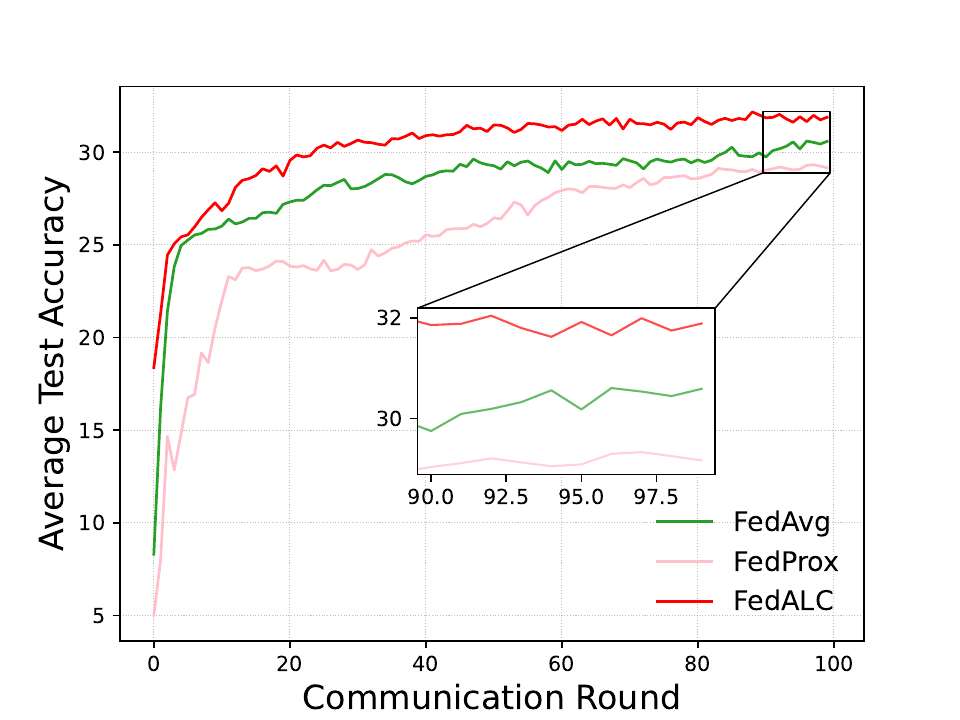}}
\caption{The top-1 average robust test accuracy (\%) of all methods on MNIST (top) and Fashion-MNIST (down) with Dir(0.1) under FGSM attack in each global round.}
\label{comm_effici_robust}
\end{figure}

\begin{figure}[!htp]
\centering
\subfigure{\includegraphics[width=0.39\textwidth]{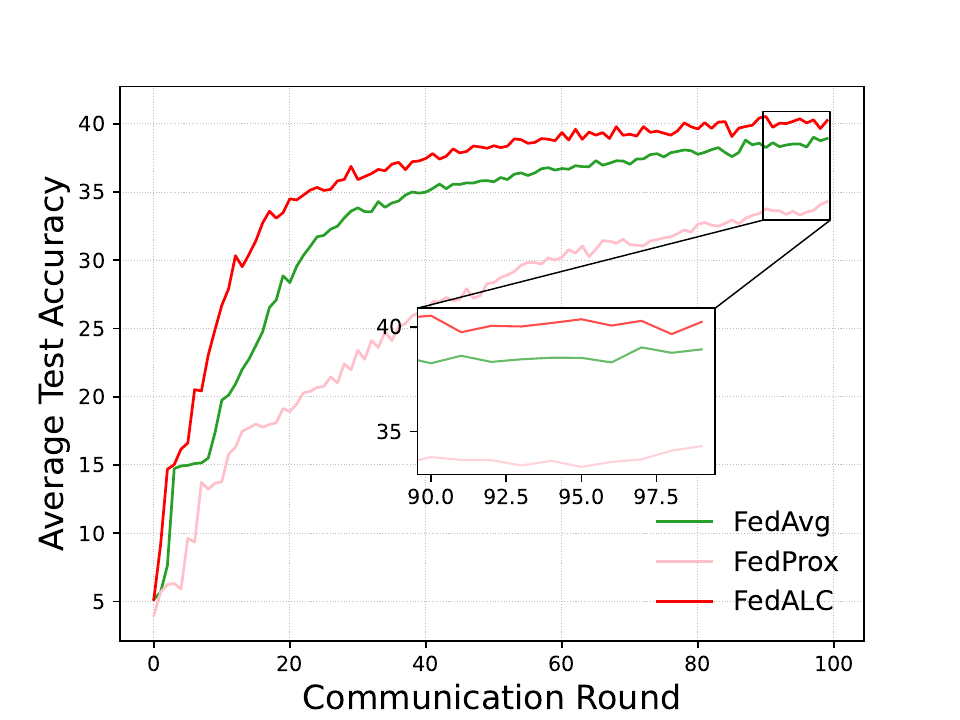}}
\subfigure{\includegraphics[width=0.39\textwidth]{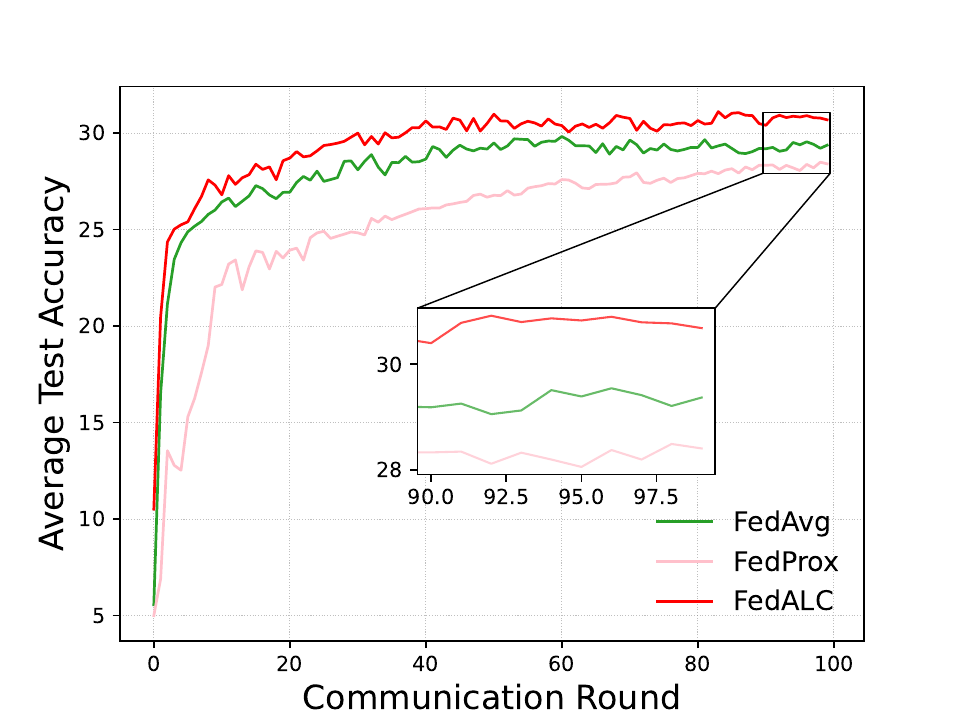}}
\caption{The top-1 average robust test accuracy (\%) of all methods on MNIST (top) and Fashion-MNIST (down) with Dir(0.1) under PGD attack in each global round.}
\label{comm_effici_robust_pgd}
\end{figure}

\begin{figure}[t]
\centering
\subfigure{\includegraphics[width=0.39\textwidth]{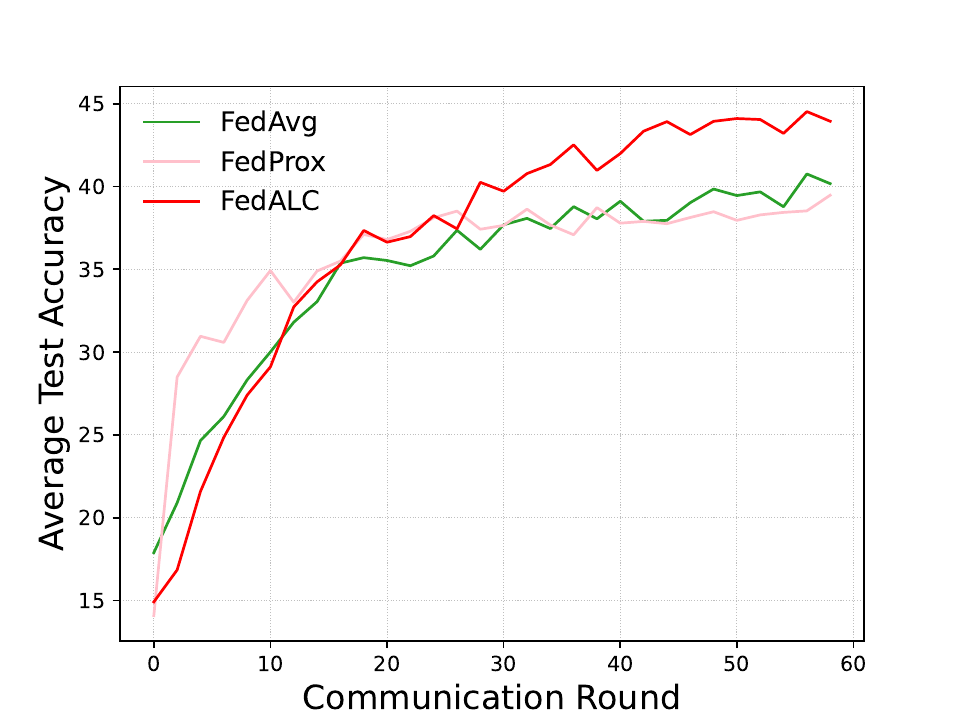}}
\caption{The top-1 average natural test accuracy (\%) of all methods on CIFAR-10 with Dir(0.1) in each global round.}
\label{comm_effici_natural_CIFAR}
\end{figure}

\begin{table}[!htp]
\centering
\caption{Natural accuracy (\%) on CIFAR-10. The best results are highlighted in \textbf{bold}.}
\label{tab:non-IID_Cifar}
\resizebox{0.49\textwidth}{!}{
\begin{tabular}{c|cccc}
\toprule
$\alpha$ & $\alpha=0.05$ & $\alpha=0.1$ & $\alpha=0.3$ & $\alpha=0.5$ \\
\midrule
FedAvg  & 32.99 $\pm$ 2.20 & 40.07 $\pm$ 2.18 & 49.67 $\pm$ 1.79 & 56.64 $\pm$ 0.68  \\
FedProx & 30.14 $\pm$ 2.24 & 34.62 $\pm$ 2.83 & \textbf{52.34} $\pm$ 1.07 & 57.34 $\pm$ 0.49  \\
FedALC (ours) & 
\textbf{35.99} $\pm$ 1.81 & \textbf{43.12} $\pm$ 1.40 & 51.50 $\pm$ 1.39 & \textbf{57.92} $\pm$ 0.69 \\
\midrule
$\alpha$ & $\alpha=0.8$ & $\alpha=1.0$ & $\alpha=2.0$ & $\alpha=5.0$ \\
\midrule
FedAvg  & 60.75 $\pm$ 0.48 & 61.95 $\pm$ 0.59 & 62.63 $\pm$ 0.58 & \textbf{64.18} $\pm$ 0.46  \\
FedProx & 59.56 $\pm$ 0.81 & 61.41 $\pm$ 0.47 & 63.46 $\pm$ 0.44 & 63.97 $\pm$ 0.66  \\
FedALC (ours) & 
\textbf{60.97} $\pm$ 0.63 & \textbf{62.24} $\pm$ 0.56 & \textbf{63.65} $\pm$ 0.41 &63.94 $\pm$ 0.58 \\
\bottomrule
\end{tabular}}
\end{table}

\subsection{Performance Comparison}
\textbf{Accuracy Comparison.} 
In our experiments, we use PyTorch to implement FedALC, FedAvg~\cite{mcmahan2017communication}, and FedProx~\cite{li2020federated}, where FedProx introduces an additional hyperparameter in the local loss function by reducing the distance between local model parameters and global model parameters to adjust local training. Here we use the hyperparameter value 0.001 reported in their work~\cite{li2020federated}. In addition, all methods are reported based on the average test accuracy over the last 10 iterations. The experimental results are reported in Table~\ref{tbl:dataset} with different levels of data heterogeneity. It appears that our proposal gains higher natural test accuracy and robust test accuracy compared to baselines in most cases.

Specifically, FedALC performs best on all metrics (natural, FGSM, BIM, CW, and PGD) except when $\alpha=0.5$. For example, when $\alpha=0.05$, FedALC achieves 85.62\% natural accuracy on MNIST and 71.84\% natural accuracy on Fashion-MNIST, which has a big advantage compared with other baselines. In contrast, the performance of the FedAvg and FedProx algorithms is relatively low in terms of both natural and robust precision. For example, under the CW attack, when $\alpha$ is 0.05, the robust accuracy of FedALC is 2.36\% and 3.23\% higher than FedAvg on MNIST and Fashion-MNIST, respectively. Similar results can also be found under other heterogeneous settings. In summary, compared with baselines, it can be found that FedALC performs well in defending against adversarial attacks in most cases, with high natural accuracy and robust accuracy. Moreover, the accuracy comparison for CIFAR-10 among different baselines is available in Table~\ref{tab:non-IID_Cifar}. Two observations can be made. First, the accuracy gradually decreases, as the heterogeneity of data distribution among devices increases. Second, our proposal outperforms other baselines in most cases.

\textbf{Communication Efficiency Comparison.} 
We present the natural test accuracies of MNIST and Fashion-MNIST under different baselines with the value $\alpha = 0.1$ in Figure~\ref{comm_effici_natural}. Moreover, the average robust test accuracies of MNIST and Fashion-MNIST, based on FGSM and PGD attack under the same heterogeneous settings, are shown in Figure~\ref{comm_effici_robust} and Figure~\ref{comm_effici_robust_pgd}, respectively. Both results indicate that our approach exhibits superior communication efficiency compared to other baselines in terms of both natural test accuracy and robust test accuracy.

More specifically, in the top subfigure in Figure~\ref{comm_effici_natural}, the natural test accuracy of FedALC is approximately 3.91\% higher than FedAvg in MNIST. Similarly, in the case of Fashion-MNIST, the natural test accuracy of FedALC in the down subfigure in Figure~\ref{comm_effici_natural} is approximately 3.49\% higher than FedAvg. However, under FGSM attack, although our performance and that of other baselines significantly decline, the results in Figure~\ref{comm_effici_robust} indicate that our approach still has an advantage over other baselines in this adversarial attack. In detail, for MNIST, its robust test accuracy is approximately 3.73\% higher than FedAvg. Similarly, for Fashion-MNIST, FedALC achieves a robust test accuracy that is approximately 4.98\% higher than FedAvg. Besides, similar observations can be found from the results in Figure~\ref{comm_effici_robust_pgd}, demonstrating that our proposal outperforms others under PGD attacks in each global communication round. Moreover, the natural test accuracy comparison for CIFAR-10 in each global round is available in Figure~\ref{comm_effici_natural_CIFAR}. The results show that although our proposal lags behind the baselines in the first 20 iterations, as the number of iterations increases, our proposal significantly surpasses the baselines.

\section{Conclusion and Future Work}
\label{sec:cons}
In this article, we have introduced a simple yet effective logits calibration adversarial learning framework for federated management, termed FedALC, to improve natural and robust accuracy under data non-IID settings. Specifically, we propose a logits calibration method, in which in each batch of model training, the logit value of each class is multiplied by the root of its corresponding inverse frequency, so as to balance the learning of the model for non-IID data distribution. Extensive experiments on three baseline datasets show that the proposed method achieves significant performance gains in both natural accuracy and robust accuracy in most cases. In the future, our method will be theoretically demonstrated and evaluated on a wider range of models to further test its effectiveness and applicability.

\bibliographystyle{IEEEtran}
\bibliography{bib_global,bib_local}

\begin{thebibliography}{10}
\providecommand{\url}[1]{#1}
\csname url@samestyle\endcsname
\providecommand{\newblock}{\relax}
\providecommand{\bibinfo}[2]{#2}
\providecommand{\BIBentrySTDinterwordspacing}{\spaceskip=0pt\relax}
\providecommand{\BIBentryALTinterwordstretchfactor}{4}
\providecommand{\BIBentryALTinterwordspacing}{\spaceskip=\fontdimen2\font plus
\BIBentryALTinterwordstretchfactor\fontdimen3\font minus
  \fontdimen4\font\relax}
\providecommand{\BIBforeignlanguage}[2]{{%
\expandafter\ifx\csname l@#1\endcsname\relax
\typeout{** WARNING: IEEEtran.bst: No hyphenation pattern has been}%
\typeout{** loaded for the language `#1'. Using the pattern for}%
\typeout{** the default language instead.}%
\else
\language=\csname l@#1\endcsname
\fi
#2}}
\providecommand{\BIBdecl}{\relax}
\BIBdecl

\bibitem{wang2023wireless}
X.~Wang, J.~Li, Z.~Ning, Q.~Song, L.~Guo, S.~Guo, and M.~S. Obaidat, ``Wireless
  powered mobile edge computing networks: A survey,'' \emph{ACM Computing
  Surveys}, 2023.

\bibitem{adhikary2023artificial_noms}
A.~Adhikary, M.~S. Munir, A.~D. Raha, Y.~Qiao, and C.~S. Hong, ``Artificial
  intelligence framework for target oriented integrated sensing and
  communication in holographic mimo,'' in \emph{NOMS 2023-2023 IEEE/IFIP
  Network Operations and Management Symposium}.\hskip 1em plus 0.5em minus
  0.4em\relax IEEE, 2023, pp. 1--7.

\bibitem{raha2023segment}
A.~D. Raha, A.~Adhikary, M.~S. Munir, Y.~Qiao, and C.~S. Hong, ``Segment
  anything model aided beam prediction for the millimeter wave communication,''
  in \emph{2023 24st Asia-Pacific Network Operations and Management Symposium
  (APNOMS)}.\hskip 1em plus 0.5em minus 0.4em\relax IEEE, 2023, pp. 113--118.

\bibitem{raha2023generative}
A.~D. Raha, M.~S. Munir, A.~Adhikary, Y.~Qiao, and C.~S. Hong, ``Generative
  ai-driven semantic communication framework for nextg wireless network,''
  \emph{arXiv preprint arXiv:2310.09021}, 2023.

\bibitem{mcmahan2017communication}
B.~McMahan, E.~Moore, D.~Ramage, S.~Hampson, and B.~A. y~Arcas,
  ``Communication-efficient learning of deep networks from decentralized
  data,'' in \emph{Artificial intelligence and statistics}.\hskip 1em plus
  0.5em minus 0.4em\relax PMLR, 2017, pp. 1273--1282.

\bibitem{qiao2023mp}
Y.~Qiao, M.~S. Munir, A.~Adhikary, H.~Q. Le, A.~D. Raha, C.~Zhang, and C.~S.
  Hong, ``Mp-fedcl: Multiprototype federated contrastive learning for edge
  intelligence,'' \emph{IEEE Internet of Things journal}, 2023.

\bibitem{zhu2021federated}
H.~Zhu, J.~Xu, S.~Liu, and Y.~Jin, ``Federated learning on non-iid data: A
  survey,'' \emph{Neurocomputing}, vol. 465, pp. 371--390, 2021.

\bibitem{qiao2023cdfed}
Y.~Qiao, M.~S. Munir, A.~Adhikary, A.~D. Raha, and C.~S. Hong, ``Cdfed:
  Contribution-based dynamic federated learning for managing system and
  statistical heterogeneity,'' in \emph{NOMS 2023-2023 IEEE/IFIP Network
  Operations and Management Symposium}.\hskip 1em plus 0.5em minus 0.4em\relax
  IEEE, 2023.

\bibitem{li2020federated}
T.~Li, A.~K. Sahu, M.~Zaheer, M.~Sanjabi, A.~Talwalkar, and V.~Smith,
  ``Federated optimization in heterogeneous networks,'' \emph{Proceedings of
  Machine learning and systems}, vol.~2, pp. 429--450, 2020.

\bibitem{zizzo2020fat}
G.~Zizzo, A.~Rawat, M.~Sinn, and B.~Buesser, ``Fat: Federated adversarial
  training,'' \emph{arXiv preprint arXiv:2012.01791}, 2020.

\bibitem{hong2021federated}
J.~Hong, H.~Wang, Z.~Wang, and J.~Zhou, ``Federated robustness propagation:
  Sharing adversarial robustness in federated learning,'' \emph{arXiv preprint
  arXiv:2106.10196}, vol.~1, 2021.

\bibitem{lyu2022privacy}
L.~Lyu, H.~Yu, X.~Ma, C.~Chen, L.~Sun, J.~Zhao, Q.~Yang, and S.~Y. Philip,
  ``Privacy and robustness in federated learning: Attacks and defenses,''
  \emph{IEEE transactions on neural networks and learning systems}, 2022.

\bibitem{goodfellow2014explaining}
I.~J. Goodfellow, J.~Shlens, and C.~Szegedy, ``Explaining and harnessing
  adversarial examples,'' \emph{arXiv preprint arXiv:1412.6572}, 2014.

\bibitem{shafahi2019adversarial}
A.~Shafahi, M.~Najibi, M.~A. Ghiasi, Z.~Xu, J.~Dickerson, C.~Studer, L.~S.
  Davis, G.~Taylor, and T.~Goldstein, ``Adversarial training for free!''
  \emph{Advances in Neural Information Processing Systems}, vol.~32, 2019.

\bibitem{madry2017towards}
A.~Madry, A.~Makelov, L.~Schmidt, D.~Tsipras, and A.~Vladu, ``Towards deep
  learning models resistant to adversarial attacks,'' \emph{arXiv preprint
  arXiv:1706.06083}, 2017.

\bibitem{zhang2023delving}
J.~Zhang, B.~Li, C.~Chen, L.~Lyu, S.~Wu, S.~Ding, and C.~Wu, ``Delving into the
  adversarial robustness of federated learning,'' \emph{arXiv preprint
  arXiv:2302.09479}, 2023.

\bibitem{chen2022calfat}
C.~Chen, Y.~Liu, X.~Ma, and L.~Lyu, ``Calfat: Calibrated federated adversarial
  training with label skewness,'' \emph{arXiv preprint arXiv:2205.14926}, 2022.

\bibitem{shah2021adversarial}
D.~Shah, P.~Dube, S.~Chakraborty, and A.~Verma, ``Adversarial training in
  communication constrained federated learning,'' \emph{arXiv preprint
  arXiv:2103.01319}, 2021.

\bibitem{qiao2023framework}
Y.~Qiao, M.~S. Munir, A.~Adhikary, A.~D. Raha, S.~H. Hong, and C.~S. Hong, ``A
  framework for multi-prototype based federated learning: Towards the edge
  intelligence,'' in \emph{2023 International Conference on Information
  Networking (ICOIN)}.\hskip 1em plus 0.5em minus 0.4em\relax IEEE, 2023, pp.
  134--139.

\bibitem{li2021model}
Q.~Li, B.~He, and D.~Song, ``Model-contrastive federated learning,'' in
  \emph{Proceedings of the IEEE/CVF Conference on Computer Vision and Pattern
  Recognition}, 2021, pp. 10\,713--10\,722.

\bibitem{qiao2023boosting}
Y.~Qiao, H.~Q. Le, and C.~S. Hong, ``Boosting federated learning convergence
  with prototype regularization,'' \emph{arXiv preprint arXiv:2307.10575},
  2023.

\bibitem{zhang2022federated}
J.~Zhang, Z.~Li, B.~Li, J.~Xu, S.~Wu, S.~Ding, and C.~Wu, ``Federated learning
  with label distribution skew via logits calibration,'' in \emph{International
  Conference on Machine Learning}.\hskip 1em plus 0.5em minus 0.4em\relax PMLR,
  2022, pp. 26\,311--26\,329.

\bibitem{menon2020long}
A.~K. Menon, S.~Jayasumana, A.~S. Rawat, H.~Jain, A.~Veit, and S.~Kumar,
  ``Long-tail learning via logit adjustment,'' \emph{arXiv preprint
  arXiv:2007.07314}, 2020.

\bibitem{lecun1998gradient}
Y.~LeCun, L.~Bottou, Y.~Bengio, and P.~Haffner, ``Gradient-based learning
  applied to document recognition,'' \emph{Proceedings of the IEEE}, vol.~86,
  no.~11, pp. 2278--2324, 1998.

\bibitem{xiao2017fashion}
H.~Xiao, K.~Rasul, and R.~Vollgraf, ``Fashion-mnist: a novel image dataset for
  benchmarking machine learning algorithms,'' \emph{arXiv preprint
  arXiv:1708.07747}, 2017.

\bibitem{krizhevsky2009learning}
A.~Krizhevsky, G.~Hinton \emph{et~al.}, ``Learning multiple layers of features
  from tiny images,'' 2009.

\bibitem{he2016deep}
K.~He, X.~Zhang, S.~Ren, and J.~Sun, ``Deep residual learning for image
  recognition,'' in \emph{Proceedings of the IEEE conference on computer vision
  and pattern recognition}, 2016, pp. 770--778.

\bibitem{mu2023fedproc}
X.~Mu, Y.~Shen, K.~Cheng, X.~Geng, J.~Fu, T.~Zhang, and Z.~Zhang, ``Fedproc:
  Prototypical contrastive federated learning on non-iid data,'' \emph{Future
  Generation Computer Systems}, vol. 143, pp. 93--104, 2023.

\bibitem{kingma2014adam}
D.~P. Kingma and J.~Ba, ``Adam: A method for stochastic optimization,''
  \emph{arXiv preprint arXiv:1412.6980}, 2014.

\bibitem{kurakin2018adversarial}
A.~Kurakin, I.~J. Goodfellow, and S.~Bengio, ``Adversarial examples in the
  physical world,'' in \emph{Artificial intelligence safety and
  security}.\hskip 1em plus 0.5em minus 0.4em\relax Chapman and Hall/CRC, 2018,
  pp. 99--112.

\bibitem{carlini2017towards}
N.~Carlini and D.~Wagner, ``Towards evaluating the robustness of neural
  networks,'' in \emph{2017 ieee symposium on security and privacy (sp)}.\hskip
  1em plus 0.5em minus 0.4em\relax Ieee, 2017, pp. 39--57.

\end{thebibliography}

\end{document}